\documentclass{article}

\usepackage{eccv26}
\usepackage[utf8]{inputenc} % allow utf-8 input
\usepackage[T1]{fontenc}    % use 8-bit T1 fonts
\usepackage{hyperref}       % hyperlinks
\usepackage{url}            % simple URL typesetting
\usepackage{booktabs}       % professional-quality tables
\usepackage{amsfonts}       % blackboard math symbols
\usepackage{nicefrac}       % compact symbols for 1/2, etc.
\usepackage{microtype}      % microtypography
\usepackage{lipsum}
\usepackage{amsmath}
\usepackage{algorithm}
\usepackage{algpseudocode}
\usepackage{multirow}
\usepackage{fancyhdr}       % header
\usepackage{graphicx}       % graphics
\graphicspath{{media/}}     % organize your images and other figures under media/ folder
\usepackage{caption}
\usepackage{subcaption}
\usepackage{listings}

\title{Puppet-CNN: Continuous Parameter Dynamics for Input-Adaptive Convolutional Networks
}

\author{
  Yucheng Xing \\
  Department of Electrical and Computer Engineering \\
  Stony Brook University \\
  Stony Brook, NY 11794, USA \\
  \texttt{yucheng.xing@stonybrook.edu} \\
  \And
  Xin Wang \\
  Department of Electrical and Computer Engineering \\
  Stony Brook University \\
  Stony Brook, NY 11794, USA \\
  \texttt{x.wang@stonybrook.edu} \\
}

\begin{document}
\maketitle

%\blfootnote{This is a preprint version of the paper under review.}

%--------------------------------------------------
%       Section 0 : Abstract
%--------------------------------------------------

%% 0.5 Page Max

\begin{abstract}

Modern convolutional neural networks (CNNs) organize computation as a discrete stack of layers whose parameters are independently stored and learned, with the number of layers fixed as an architectural hyperparameter. In this work, we explore an alternative perspective: \textit{can network parameterization itself be modeled as a continuous dynamical system?} We introduce \textit{\textbf{Puppet-CNN}}, a framework that represents convolutional layer parameters as states evolving along a learned parameter flow governed by a neural ordinary differential equation (ODE). Under this formulation, layer parameters are generated through continuous evolution in parameter space, and the effective number of generated layers is determined by the integration horizon of the learned dynamics, which can be modulated by input complexity to enable input-adaptive computation. We validate this formulation on standard image classification benchmarks and demonstrate that continuous parameter dynamics can achieve competitive predictive performance while substantially reducing stored trainable parameters. These results suggest that viewing neural network parameterization through the lens of dynamical systems provides a structured and flexible design space for adaptive convolutional models. 

\end{abstract}

%--------------------------------------------------
%       Section 1 : Introduction
%--------------------------------------------------

%% 1 Page Max

\section{Introduction~\label{sec:1}}

Modern convolutional neural networks (CNNs) organize computation as a discrete stack of layers whose parameters are independently stored and learned. In this formulation, the number of layers is fixed as an architectural hyperparameter, and layer parameters are treated as separate tensors across depth. This discrete organization has driven much of the progress in modern vision models. However, it also implicitly assumes that parameterization across depth is static rather than structured as a generative process. 

In this work, we explore an alternative perspective: \textit{can layer parameterization itself be modeled as a continuous dynamical system?} Instead of specifying each layer's parameters independently, we propose to organize convolutional kernels as states evolving along a learned trajectory in parameter space. Under this view, depth is interpreted through the integration horizon of an underlying dynamical process that sequentially generates layer parameters, rather than as a pre-defined stack of fixed tensors. 

We instantiate this idea through \textit{\textbf{Puppet-CNN}}, a framework consisting of two components: a compact dynamical generator (the \textit{puppeteer}) and a standard convolutional backbone (the \textit{puppet}). The puppeteer is formulated as a neural ordinary differential equation (ODE) that governs the continuous evolution of convolutional parameters. By integrating this learned parameter flow, Puppet-CNN generates the kernels of successive layers through a shared dynamical mechanism, thereby introducing structured coupling across depth. While other sequential models could in principle parameterize depth-wise evolution, the continuous-time formulation provides a direct interpretation in which the effective number of generated layers corresponds to the integration interval of the underlying dynamics. 

In many real-world scenarios, input samples exhibit heterogeneous levels of structural complexity and may benefit from different amounts of computational processing. Conventional CNNs, however, apply a fixed-depth architecture uniformly across inputs. Under the proposed dynamical formulation, depth corresponds to the integration horizon, allowing computation to vary by modulating the integration process. As a result, both network structure and parameters are generated jointly within a unified continuous framework, rather than being selected or pruned from a fixed pre-trained architecture.

We evaluate Puppet-CNN on standard image classification benchmarks to validate the feasibility of this formulation. Our results demonstrate that modeling parameterization as continuous dynamics can preserve competitive predictive performance while maintaining a compact parameterization. These findings suggest that viewing neural network parameterization through the lens of dynamical systems provides a structured and flexible design space for adaptive convolutional architectures. 

The contributions of this paper are three-fold:
\begin{itemize}
    \item We propose a continuous parameter dynamics formulation for convolutional neural networks, in which layer parameters are modeled as states evolving along a learned trajectory governed by a neural ordinary differential equation.
    \item We reinterpret network depth as the integration horizon of the underlying parameter dynamics, enabling a unified mechanism that jointly generates both network structure and layer parameters. 
    \item We show that input-adaptive computation emerges naturally from this dynamical formulation by modulating the integration process, and validate the feasibility of this perspective on standard image classification benchmarks. 
\end{itemize}

The organization of the remainder of the paper is as follows: In Sec.~\ref{sec:2}, we give an overview of the related works. We introduce our model in detail in Sec.~\ref{sec:3} and demonstrate its effectiveness through experiments in Sec.~\ref{sec:4}. Finally, we conclude our work in Sec.~\ref{sec:5}. Code will be released upon acceptance.

%--------------------------------------------------
%       Section 2 : Related Works
%--------------------------------------------------

%% 1 Page Max

\section{Related Works~\label{sec:2}}

%\vspace{-0.1in}
\subsection{Depth-Variant Computation~\label{sec:2.1}}
%\vspace{-0.05in}
Increasing network depth and width has been widely associated with improved representational capacity in convolutional neural networks~\cite{He_2016_CVPR, NIPS2012_c399862d, 7780677}. At the same time, it has been observed that different inputs may require different amounts of computational processing~\cite{https://doi.org/10.48550/arxiv.1605.06431}. This observation has motivated a line of research on depth-variant or adaptive computation mechanisms. Early-Exiting methods~\cite{10.5555/3305381.3305436, 10.5555/3304889.3304963, https://doi.org/10.48550/arxiv.1709.01686, https://doi.org/10.48550/arxiv.1706.00885} allow samples to exit a deep model at intermediate layers based on pre-defined criteria, such as intermediate classification confidence or uncertainty. These approaches typically operate on standard CNN backbones, sometimes augmented with multi-resolution branches or densely connected structures as in~\cite{huang2018multiscale, 10.1145/3340531.3411973, Yang_2020_CVPR}. Other strategies modify execution paths within a fixed architecture. For example, Layer Skipping methods~\cite{Wang_2018_ECCV, Veit2018} use gating mechanisms to decide whether certain layers should be executed. Similarly, approaches~\cite{Figurnov_2017_CVPR, leroux2018iamnn:, 8954249} dynamically adjust the number of residual transformations applied within each block. These methods share a common characteristic that adaptivity is realized by selecting, skipping, or reusing components from a pre-defined network structure. The underlying layer parameters are learned and stored explicitly, while variation in computation arises from conditional activation within the existing architecture.

%\vspace{-0.15in}
\subsection{Input-Conditioned Parameterization~\label{sec:2.2}}
%\vspace{-0.05in}
In conventional CNNs, parameters are optimized over the training set and then fixed for all inputs during inference. To introduce input-dependent behavior, a line of work explores input-conditioned parameterization, where convolutional kernels are adapted or generated as functions of the input~\cite{6796337}. One group of approaches adjusts a fixed set of base parameters through input-conditioned modulation. For example, CondConv~\cite{NEURIPS2019_f2201f51} combines multiple expert kernels using sample-dependent routing weights, and Dynamic Convolution~\cite{Chen_2020_CVPR} further constrains the normalized combination coefficients. Related designs incorporate spatially varying attention mechanisms to modulate convolutional responses according to local content~\cite{Harley_2017_ICCV, Su_2019_CVPR}. In these methods, the kernel tensors are pre-defined, while input-dependent coefficients determine how they are activated for each sample. Another group of approaches directly generates parameters from the input using an auxiliary network. Early work~\cite{Denil2013PredictingPI} factorized parameter matrices and predicted part of the factors at test time. Dynamic Filter Networks (DFNs)~\cite{debrabandere16dynamic} and HyperNetworks~\cite{ha2017hypernetworks} generate layer parameters on the fly through a separate model. Subsequent extensions explore spatial-specific generation~\cite{inbook_LS_DFN}, decoupled dynamic kernels~\cite{Zhou_2021_CVPR}, meta-learned generators~\cite{https://doi.org/10.48550/arxiv.1703.00837}, and grouped transformations as in WeightNet~\cite{ma2020weightnet}. Input-conditioned generation has also been combined with additional contextual signals such as camera perspective~\cite{NIPS2017_e7e23670} or graph edge attributes~\cite{Simonovsky_2017_CVPR}. Despite differences in implementation, these approaches share a common formulation in which parameter adaptation or generation is performed in a layer-wise manner. Parameters for each layer are produced or modulated through mappings conditioned on the corresponding layer-wise input features, while the organization of parameters across depth remains discretely defined by the underlying architecture. 

%\vspace{-0.15in}
\subsection{Parameter Generation Paradigms~\label{sec:2.3}}
%\vspace{-0.05in}
Beyond input-conditioned parameterization, prior work reflects different paradigms for organizing generated parameters at the network level. In layer-wise generation approaches~\cite{debrabandere16dynamic, ha2017hypernetworks, inbook_LS_DFN, Zhou_2021_CVPR, ma2020weightnet}, parameters for successive layers are instantiated independently through separate mappings. Although such parameters may be conditioned on intermediate features, each layer’s weights are generated as a discrete entity, and the depth of the network remains pre-defined. A different paradigm synthesizes parameters at the global level. Diffusion-based approaches~\cite{wang2024neural, soro2024diffusion} treat the entire parameter set as a sample from a learned distribution and generate a complete configuration for a fixed architecture. In the reported experiments, parameter synthesis is typically demonstrated on relatively small networks or on subsets of layers, where the parameter space can be represented explicitly as a generative target. In these methods, the generative ``time'' corresponds to the sampling process in parameter space rather than to the depth dimension of the network, and the structural organization of layers remains unchanged. Related ideas also appear in meta-learning frameworks~\cite{https://doi.org/10.48550/arxiv.1703.00837}, where parameter generators or meta-learners adapt model weights across tasks. In such settings, adaptation typically occurs at the task level, producing task-specific parameter configurations while preserving a pre-defined network structure. Taken together, existing approaches reflect two dominant parameter generation paradigms: layer-wise parameter mapping and global parameter synthesis under fixed architectures. In both cases, network depth is treated as a pre-defined structural dimension. Generated parameters are either mapped independently per layer or sampled as a complete set, rather than organized as states evolving along depth under a shared dynamical mechanism.

%--------------------------------------------------
%       Section 3 : Methodology
%--------------------------------------------------

% 2 Pages Max

\section{Methodology~\label{sec:3}}

\begin{figure*}[!htpb]
    %\vspace{-0.1in}
    \centering
    \includegraphics[width=0.7\linewidth]{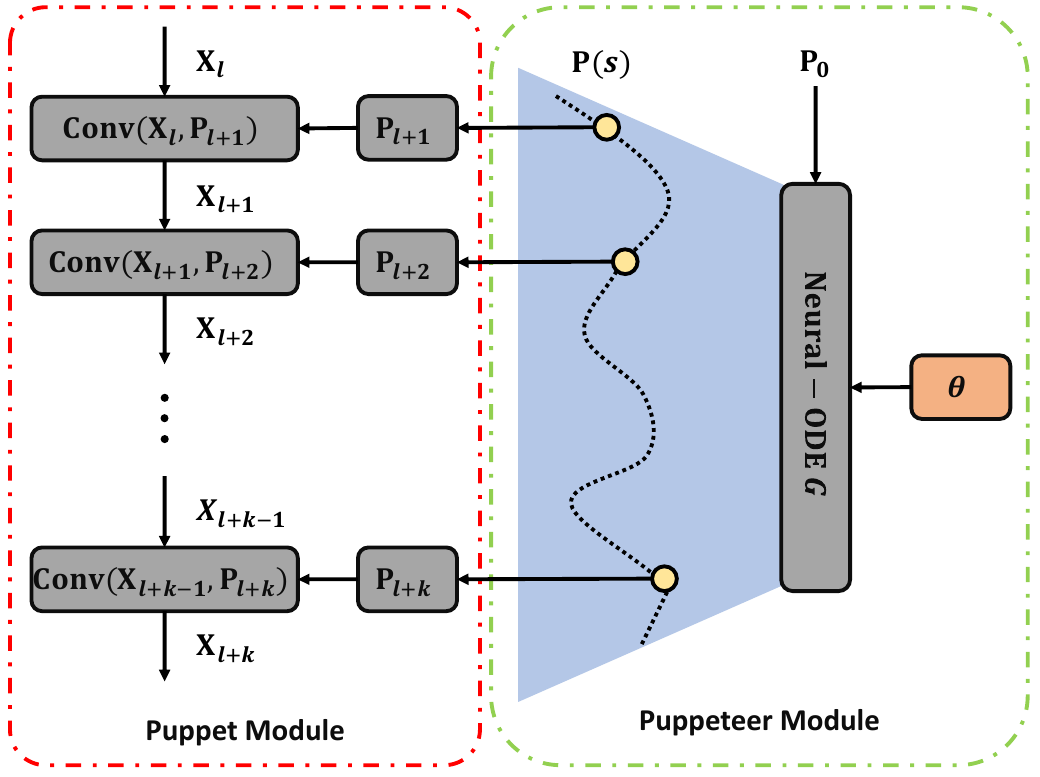}
    %\vspace{-0.1in}
    \caption{Overview of Puppet-CNN. The puppeteer module governs continuous parameter evolution, and the generated parameters define the layers of the puppet network.}
    \label{fig:framework_new}
    %\vspace{-0.2in}
\end{figure*}

This section presents the overall architecture of \textit{Puppet-CNN} under the continuous parameter dynamics perspective. The framework consists of two components: a \textit{puppeteer} module, which generates convolutional parameters through a continuous evolution process, and a \textit{puppet} module, which applies the generated parameters to process input data (see Fig.~\ref{fig:framework_new}). Instead of assigning independent parameters to each layer, the puppeteer organizes parameterization across depth as a continuous dynamical system in parameter space, so that convolutional layers are obtained by discretizing a shared continuous parameter trajectory. In addition, the evolution process can be modulated according to input characteristics, enabling adaptive computation. The following subsections formalize this formulation and describe its practical instantiation. 

\subsection{Continuous Parameter Evolution~\label{sec:3.1}}

We first formalize the continuous parameter evolution underlying Puppet-CNN. Let $s \in [0, 1]$ denote a normalized continuous evolution coordinate, and let $\mathbf{P}(s)$ represent the convolutional kernel parameters as a function of $s$. Rather than assigning independent parameter tensors to discrete layers, we model their variation along the evolution coordinate as a continuous process governed by an ordinary differential equation (ODE):
%\vspace{-0.05in}
\begin{equation}
    %\small
    \frac{d\mathbf{P}(s)}{ds} = G(\mathbf{P}(s); \theta),~\label{eq:ode_derivative}
%\vspace{-0.05in}
\end{equation}
where $G(\cdot; \theta)$ is a learnable neural function parameterized by $\theta$ that specifies the rate of change of parameters with respect to $s$. This formulation induces a continuous trajectory in parameter space, where parameters are interpreted as states evolving under a shared transformation rule.

To instantiate a finite network from the continuous formulation, we consider the solution of Eq.~\ref{eq:ode_derivative} over a small interval $[s, s+\Delta s]$, which can be expressed in integral form as
%\vspace{-0.05in}
\begin{equation}
    %\small
    \mathbf{P}(s + \Delta s) = \mathbf{P}(s) + \int^{s + \Delta s}_{s}G(\mathbf{P}(\tau); \theta)\, d\tau.~\label{eq:ode_integral}
%\vspace{-0.05in}
\end{equation}
Since this integral generally does not admit a closed-form solution, we approximate it numerically. Using an explicit Euler scheme with step size $\Delta s$, the evolution can be discretized as
%\vspace{-0.05in}
\begin{equation}
    %\small
    \mathbf{P}_{l + 1} = \mathbf{P}_{l} + G(\mathbf{P}_{l}; \theta)\,\Delta s,~\label{eq:ode_propagation} 
%%\vspace{-0.05in}
\end{equation}
where $\mathbf{P}_{l} \approx \mathbf{P}(s_{l})$ and $s_{l} = l\,\Delta s$. Each propagation step yields one instantiated convolutional transformation, whose kernel is derived from the sampled state $\mathbf{P}_{l}$. Under the normalized evolution interval $[0, 1]$, the effective depth of the network is determined by the sampling resolution, i.e.,
%\vspace{-0.05in}
\begin{equation}
    %\small
    D = \lfloor\frac{1}{\Delta s}\rfloor,~\label{eq:depth}
%\vspace{-0.05in}
\end{equation}
so that network depth arises from discretizing a single underlying evolution trajectory rather than from pre-specified discrete layers. 

%\begin{figure}[!htpb]
%    %%\vspace{-0.1in}
%    \centering
%    \includegraphics[width=\linewidth]{fig/weight_gen.pdf}
%    %\vspace{-0.1in}
%    \caption{The illustration of the weight generation process.}
%    \label{fig:weight_gen}
%    %\vspace{-0.2in}
%\end{figure}

\subsection{Input-Adaptive Parameter Trajectory Sampling~\label{sec:3.2}}

The continuous parameter evolution framework naturally enables input-adaptive computation. In practical scenarios, input samples exhibit heterogeneous structural complexity and may benefit from different amounts of processing. Within our formulation, adaptivity arises by modulating the sampling of a shared continuous parameter trajectory for each input. Specifically, both the initial condition of the trajectory and the discretization resolution along it can depend on input characteristics. As a result, Puppet-CNN supports adaptive behavior at two complementary levels: \textit{parameter-level adaptation} through trajectory initialization and \textit{depth-level adaptation} through trajectory sampling density.

\noindent\textbf{Parameter-Level Adaptation.}
As indicated by Eq.~\ref{eq:ode_integral}, the underlying continuous trajectory $\mathbf{P}(s)$ over the normalized interval $[0, 1]$ is uniquely determined by the shared evolution function $G(\cdot; \theta)$ and the initial state $\mathbf{P}_{0} = \mathbf{P}(0)$, while the discretized samples used for network instantiation depend on the chosen step size. By allowing this initial condition to depend on a scalar complexity signal $c(\mathbf{X}_{0})$ extracted from the input sample $\mathbf{X}_{0}$, 
%\vspace{-0.05in}
\begin{equation}
    %\small
    \mathbf{P}_{0} = \psi(c(\mathbf{X}_{0})),~\label{eq:initial_adapt}
%\vspace{-0.05in}
\end{equation}
where $\psi(\cdot)$ denotes a lightweight mapping from the scalar complexity measure to the initial parameter state, different inputs induce distinct parameter trajectories under the same evolution rule. 

\noindent\textbf{Depth-Level Adaptation.}
As discussed in Sec.~\ref{sec:3.1}, the effective depth of the instantiated network is determined by the discretization step size $\Delta s$. To enable input-dependent depth variation, we allow the step size to depend on the same complexity signal $c(\mathbf{X}_{0})$, 
%\vspace{-0.05in}
\begin{equation}
    %\small
    \Delta s = \phi(c(\mathbf{X}_{0})),~\label{eq:depth_adapt}
%%\vspace{-0.05in}
\end{equation}
where $\phi(\cdot)$ is a scalar mapping that preserves a monotonic relation between the input complexity and the sampling resolution. Through this modulation, inputs with higher complexity lead to finer sampling of the shared parameter trajectory, and thus deeper instantiated networks, while simpler inputs result in coarser sampling and shallower architectures. 

These two mechanisms operate on different aspects of the same continuous parameter evolution. The evolution function $G$ remains shared across all inputs, while input-dependent variation enters only through the initial state and sampling resolution of a single underlying trajectory. Adaptive computation is therefore not introduced as an additional control mechanism, but emerges intrinsically from the continuous organization of parameters across depth. Adaptation becomes a structural consequence of parameter evolution rather than an externally imposed architectural modification.

In practice, we instantiate the scalar complexity signal $c(\mathbf{X}_{0})$ using an entropy-based measure~\cite{larkin2016reflections, wu2013local, vila2014analysis} that combines spatial- and frequency-domain statistics of the input. Specifically, we define
%\vspace{-0.05in}
\begin{equation}
    %\small
    E(\mathbf{X}_{0}) = -\sum_{x_{i}}p(x_{i})\log p(x_{i}),~\label{eq:info_entropy}
%\vspace{-0.05in}
\end{equation}
where $p(x_{i})$ denotes the empirical distribution of pixel intensities in the input. To further account for structural variation in the frequency domain, we compute the entropy of the Fourier-transformed representation $\mathcal{F}(\mathbf{X}_{0})$. The final complexity signal is defined as 
%\vspace{-0.05in}
\begin{equation}
    %\small
    c(\mathbf{X}_{0}) = \frac{1}{2}E(\mathbf{X}_{0}) + \frac{1}{2}E(\mathcal{F}(\mathbf{X}_{0})).~\label{eq:complexity}
%\vspace{-0.05in}
\end{equation}
This provides a lightweight proxy for structural complexity. The mappings $\psi(\cdot)$ and $\phi(\cdot)$ are then implemented as simple scalar transformations to generate the initial parameter state and sampling resolution, respectively. Importantly, the overall formulation is independent of the specific complexity metric, and alternative scalar measures can be incorporated within the same continuous parameter evolution framework without altering the shared evolution dynamics.

%\vspace{-0.1in}
\subsection{Puppet-Puppeteer Architecture~\label{sec:3.3}}
\begin{figure*}[!htpb]
    %\vspace{-0.2in}
    \centering
    \includegraphics[width=0.9\linewidth]{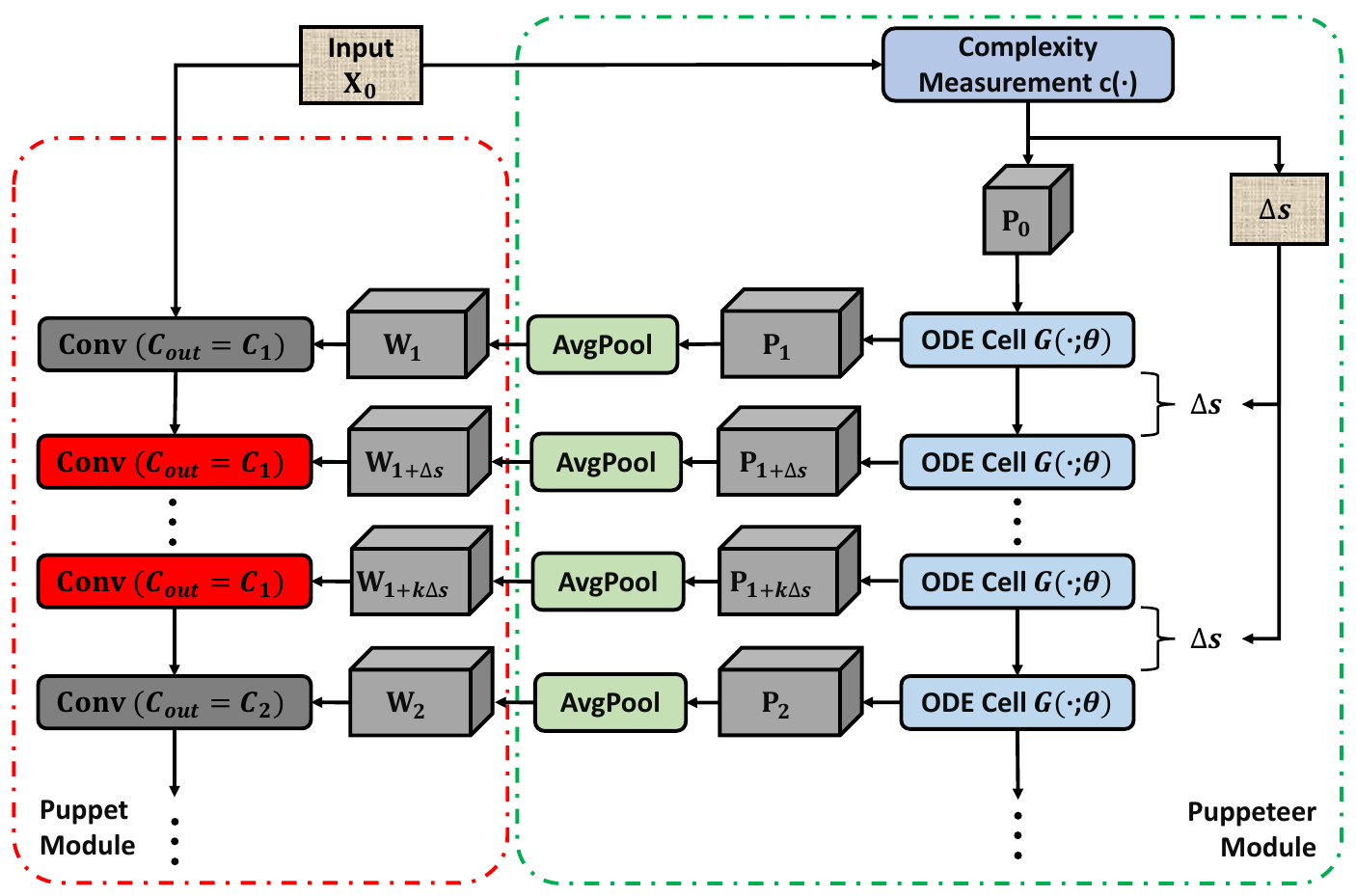}
    %\vspace{-0.1in}
    \caption{Architectural instantiation of the Puppet–Puppeteer framework. The puppeteer module evolves parameter states through an ODE trajectory and generates discretized kernel states, which are resized and injected into successive convolutional layers of the puppet network. The trajectory initialization and discretization step are determined by the input complexity measurement.}
    \label{fig:final_architecture}
    %\vspace{-0.2in}
\end{figure*}

%\vspace{-0.05in}
We now detail how the continuous parameter evolution framework described in \ref{sec:3.1} - \ref{sec:3.2} is instantiated within the Puppet–Puppeteer architecture, as illustrated in Fig.~\ref{fig:final_architecture}. In this realization, the \textit{puppeteer} module governs parameter dynamics along a continuous evolution coordinate, while the \textit{puppet} module is obtained by sampling this trajectory into discretized kernel states, which are injected into successive convolutional layers. Under this view, architectural depth is determined by how densely the trajectory is sampled, while input-dependent modulation can alter both the sampled parameters and the resulting depth.

\begin{figure*}[!htpb]
    %\vspace{-0.25in}
    \centering
    \includegraphics[width=\linewidth]{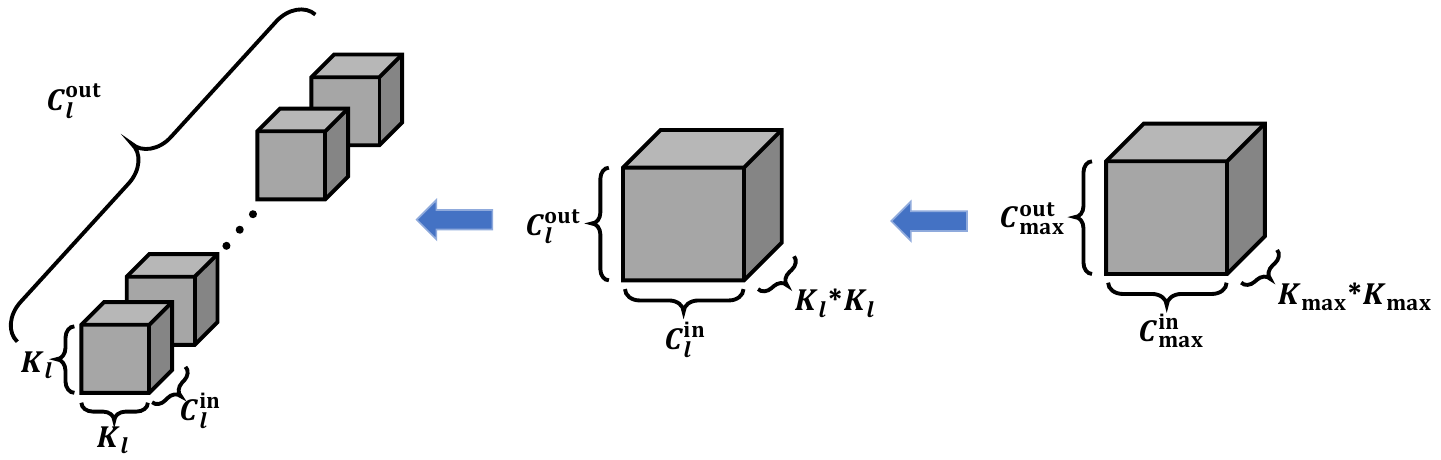}
    %\vspace{-0.25in}
    \caption{Organization of the parameter state for continuous evolution and layer-wise instantiation. The continuous evolution is performed in a fixed-dimensional maximal tensor space, where spatial kernel dimensions are reshaped for processing within the dynamical function $G(\cdot; \theta)$. Sampled states are subsequently projected to match the layer-specific kernel configurations.}
    \label{fig:kernel_resize}
    %\vspace{-0.2in}
\end{figure*}

Starting from an input-dependent initialization $\mathbf{P}_{0}$ with Eq.~\ref{eq:initial_adapt}, the puppeteer propagates parameters along the normalized evolution coordinate $s \in [0, 1]$ according to the shared dynamical function $G(\cdot; \theta)$. The continuous state $\mathbf{P}(s)$ is defined over a maximal tensor shape $(C_{\text{max}}^{\text{out}},\, C_{\text{max}}^{\text{in}},\, K_{\text{max}},\, K_{\text{max}})$, where each dimension corresponds to the largest channel or kernel configuration within the puppet module. This ensures that parameter evolution is performed within a fixed-dimensional state space. For processing within the dynamical function $G(\cdot; \theta)$, the spatial kernel dimensions are reorganized into a convolution-compatible representation, as illustrated in Fig.~\ref{fig:kernel_resize}, while preserving a unified parameter state. The discretized states $\mathbf{P}_{l}$ are obtained by sampling the evolution trajectory as defined in Eq.~\ref{eq:ode_propagation}. For each convolutional layer $l$ within the puppet module, the sampled state is projected to the required kernel dimension $(C_{l}^{\text{out}},\, C_{l}^{\text{in}},\, K_{l},\, K_{l})$ through a deterministic resizing operation. This design accommodates heterogeneous layer configurations while preserving a single continuous evolution trajectory. 

As formalized in Sec.~\ref{sec:3.1} (see Eq.~\ref{eq:depth}), discretizing the evolution interval with step size $\Delta s$ yields a finite set of sampled states $\{\mathbf{P}_{l}\}$ whose cardinality determines the effective depth $D$ of the network. Within the Puppet–Puppeteer architecture, each propagation step along the discretized trajectory gives rise to one convolutional transformation in the puppet module, with kernels derived from the corresponding sampled states. Depth therefore emerges from the sampling density of a single continuous trajectory, rather than from pre-defined discrete layer allocation. When $\Delta s$ is modulated as described in Sec.~\ref{sec:3.2}, this mechanism naturally enables input-adaptive depth variation.

Bringing the above components together, the parameter generation and feature transformation at layer index $l$ follow a unified forward process. Starting from the input-dependent initialization defined in Eq.~\ref{eq:initial_adapt},
%\vspace{-0.05in}
\begin{equation}
    %\small
    \mathbf{P}_{0} = \psi(c(\mathbf{X}_{0})),
%\vspace{-0.05in}
\end{equation}
the puppeteer module propagates parameters along the discretized evolution trajectory as defined in Eq.~\ref{eq:ode_propagation},
%\vspace{-0.05in}
\begin{equation}
    %\small
    \mathbf{P}_{l} = \mathbf{P}_{l - 1} + G(\mathbf{P}_{l - 1}; \theta)\,\Delta s,
%\vspace{-0.05in}
\end{equation}
where the step size $\Delta s$ may depend on the same complexity signal as described in Eq.~\ref{eq:depth_adapt}, 
%\vspace{-0.05in}
\begin{equation}
    %\small
    \Delta s = \phi(c(\mathbf{X}_{0})).
%\vspace{-0.05in}
\end{equation}
For each layer $l$ in the puppet module, the sampled state $\mathbf{P}_{l}$ is resized to match the required kernel dimension $(C_{l}^{\text{out}},\, C_{l}^{\text{in}},\, K_{l},\, K_{l})$, producing the convolutional kernel $\mathbf{W}_{l}$. The puppet module then applies this kernel as the convolutional transformation at layer $l$ to update the feature representation,
%\vspace{-0.05in}
\begin{equation}
    %\small
    \mathbf{X}_{l + 1} = \text{Conv}(\mathbf{X}_{l}, \mathbf{W}_{l}).
%\vspace{-0.05in}
\end{equation}
Through this unified process, continuous parameter evolution, trajectory sampling, and discrete convolutional computation are coupled in a single dynamical mechanism, from which both parameter values and architectural depth emerge. Furthermore, since all convolutional kernels are generated from the shared dynamical function $G(\cdot; \theta)$, training the puppet network implicitly reduces to optimizing the parameters $\theta$ of the puppeteer through backpropagation along the discretized trajectory.

%--------------------------------------------------
%       Section 4 : Experiments
%--------------------------------------------------

%% 3 Pages Max
%\vspace{-0.05in}
\section{Experiments~\label{sec:4}}
%\vspace{-0.05in}
We evaluate the proposed Puppet-CNN framework on image classification to assess its effectiveness in practical convolutional architectures. Although the formulation is task-agnostic and applies to general convolutional models, classification provides a controlled setting to examine parameter efficiency, adaptive depth behavior, and overall predictive performance. Through these experiments, we analyze whether organizing convolutional parameters via continuous evolution can reduce parameter redundancy while maintaining competitive predictive performance under heterogeneous input complexity.

%\vspace{-0.05in}
\subsection{Datasets~\label{sec:4.1}}
%\vspace{-0.05in}
\noindent\textbf{CIFAR-10~\cite{cifar10}:}
CIFAR-10 consists of 60,000 RGB images of size $32\times32$ across 10 categories, with 50,000 training and 10,000 testing samples. In our experiments, we randomly select 10,000 images from the official training set for model training to evaluate performance under limited data settings. From this subset, $20\%$ of the samples are further used for validation, and the remaining samples are used for training.

\noindent\textbf{CIFAR-100~\cite{cifar100}:}
CIFAR-100 contains 60,000 RGB images of size $32\times32$ across 100 categories, with 50,000 training and 10,000 testing samples. We follow the standard split and use the full training set, where $20\%$ of the training data is held out as validation for hyperparameter selection.

\noindent\textbf{mini-ImageNet~\cite{NIPS2016_90e13578}:}
mini-ImageNet is a subset of ImageNet containing 60,000 images across 100 categories. We split the dataset into training, validation, and testing sets with a ratio of 8:1:1 using a fixed random seed. All images are resized and randomly cropped to $64\times64$ during preprocessing.

%\vspace{-0.05in}
\subsection{Implementation Details~\label{sec:4.2}}
%\vspace{-0.05in}
All models are implemented in PyTorch and trained from scratch on a single NVIDIA GeForce RTX 4090 GPU. To ensure fair comparison, all baseline methods are re-implemented and trained under identical data splits and optimization settings. For CIFAR-10, 10,000 training samples are used as described in Sec.~\ref{sec:4.1}. All models are trained for $800$ epochs using the Adam optimizer with a batch size of $64$. The learning rate is initialized at $1\times10^{-3}$ and decayed to $1\times10^{-5}$ by the end of training. To ensure comparable model capacity across all methods, we restrict the maximum number of output channels to $512$ for both Puppet-CNN and all baseline models. No additional data augmentation is applied unless explicitly specified.

For Puppet-CNN, we adopt a convolutional backbone with channel configurations $\{64,128,256,512\}$. Each convolutional layer is followed by batch normalization and ReLU activation. When the channel dimension changes, a 2D max-pooling layer is applied. All convolutional kernels are of size $3\times3$ and are generated from discretized parameter states produced by the ODE propagation defined in Eq.~\ref{eq:ode_propagation}. Sampled parameter states are resized via 3D average pooling to match the required kernel dimensions before instantiation. The adaptive discretization step size is implemented as
\begin{equation}
    %\small
    \Delta s = \phi(c(\mathbf{X}_0)) = \tanh\!\left(c(\mathbf{X}_0)^{-1}\right),
\end{equation}
where $c(\mathbf{X}_0)$ denotes the entropy-based complexity defined in Eq.~\ref{eq:complexity}. The effective depth is therefore determined by $D=\lfloor 1/\Delta s \rfloor$. The initial parameter state is initialized as
\begin{equation}
    %\small
    \mathbf{P}_0 = \psi(c(\mathbf{X}_0)) = \exp\!\left(1 - c(\mathbf{X}_0)^{-2}\right),
\end{equation}
which is expanded to match the maximal kernel tensor dimension. These mappings are deterministic and introduce no additional learnable parameters.

\subsection{Overall Performance~\label{sec:4.3}}

We evaluate Puppet-CNN on CIFAR-10 and compare it with representative adaptive-parameter methods, including DFN~\cite{debrabandere16dynamic}, WeightNet~\cite{ma2020weightnet}, and DDFN~\cite{Zhou_2021_CVPR} (modified to sample-wise, denoted as DDFN-SW), as well as adaptive-depth methods such as BranchyNet~\cite{https://doi.org/10.48550/arxiv.1709.01686}, SkipNet~\cite{Wang_2018_ECCV}, and DRNN~\cite{8954249}. All models are trained under the same settings described in Sec.~\ref{sec:4.2}.
\begin{table*}[!htpb]
    %\vspace{-0.2in}
    \centering
    \renewcommand\arraystretch{0.8}	
    %\small
    \caption{Overall comparison among different models.}% on Cifar-10.}
    \label{tab:overall_cifar_cls}
    %\vspace{-0.1in}
    \begin{tabular}{c | c c c c c}
        \toprule
         \multicolumn{2}{c}{} &Top-1 Acc ($\uparrow$) &Top-5 Acc ($\uparrow$) &Params ($\downarrow$) &Speed ($\downarrow$) \\
         \midrule
         %Fixed CNN \\
         %\cmidrule{1-1}
         %AlexNet &65.86$\%$ &92.05$\%$ &11.65MB &0.0006s/img \\ 
         %VGG-11 &67.84$\%$ &91.98$\%$ &39.01MB &0.0010s/img \\ 
         %ResNet-18 &68.94$\%$ &95.17$\%$ &44.84MB &0.0019s/img \\ 
         %\midrule
         \multirow{3}{*}{\parbox{1.4 cm}{\centering Adaptive Params}} % \\
         %\cmidrule{1-1}
         &DFN &68.59$\%$ &93.13$\%$ &75.89 &0.0012 \\ 
         &WeightNet &62.77$\%$ &93.52$\%$ &45.87 &0.0040 \\ 
         &DDFN-SW &55.55$\%$ &93.34$\%$ &300.83 &0.0059 \\ 
         \midrule
         \multirow{3}{*}{\parbox{1.4 cm}{\centering Adaptive Depth}} % \\
         %\cmidrule{1-1}
         &BranchyNet &70.00$\%$ &93.94$\%$ &27.69 &0.0015 \\ 
         &SkipNet &55.82$\%$ &87.95$\%$ &68.88 &0.0115 \\
         &DRNN &41.71$\%$ &63.89$\%$ &45.78 &0.0025 \\ 
         \midrule
         \textbf{Ours} % \\
         %\cmidrule{1-1}
         &\textbf{Puppet-CNN} &\textbf{72.51$\%$} &\textbf{96.85$\%$} &\textbf{1.08} &0.0039 \\
         \bottomrule
    \end{tabular}
    %\vspace{-0.2in}
\end{table*}
Performance is evaluated using Top-1 and Top-5 accuracy ($\uparrow$), while efficiency is measured by parameter size (Params (MB), $\downarrow$) and inference speed (Speed (s/img), $\downarrow$). As shown in Table.~\ref{tab:overall_cifar_cls}, Puppet-CNN attains strong classification accuracy in this setting, while using only $1.08$\,MB of parameters, substantially fewer than the compared adaptive architectures. Overall, the results suggest that organizing convolutional parameters through continuous evolution can yield a compact model that remains competitive in predictive performance, while supporting sample-wise parameter and depth adaptation within a unified framework.

\subsection{Ablation Study~\label{sec:4.4}}

To better understand the contribution of different components in our model, we conduct ablation studies from three aspects: the \textit{Puppet-Puppeteer parameterization scheme}, \textit{parameter-level adaptation}, and \textit{depth-level adaptation}. The goal of these experiments is not to demonstrate uniformly higher accuracy, but to examine whether continuous parameter dynamics can serve as a practical and compact parameterization mechanism for convolutional neural networks.

\noindent\textbf{Puppet-Puppeteer Scheme.}
We first examine whether organizing convolutional parameters through a continuous evolution process can serve as a practical parameterization mechanism for CNN models.
To this end, we apply the proposed Puppet-Puppeteer framework to several representative CNN backbones, including AlexNet~\cite{NIPS2012_c399862d}, VGG~\cite{https://doi.org/10.48550/arxiv.1409.1556}, and ResNet~\cite{He_2016_CVPR}. 
These models represent several fundamental design patterns widely used in convolutional networks, including shallow convolutional architectures, deep sequential convolutional structures, and residual connection-based networks.
\begin{table*}[!htpb]
    %\vspace{-0.2in}
    \centering
    \renewcommand\arraystretch{0.8}	
    %\small
    \caption{Progressive ablation study on representative CNN backbones with the Puppet-Puppeteer scheme and parameter adaptation.}% on Cifar-10.}
    \label{tab:compression_cifar_cls}
    %\vspace{-0.1in}
    %\begin{tabular}{c p{0.06\linewidth} p{0.06\linewidth} p{0.1\linewidth} p{0.06\linewidth} p{0.06\linewidth} p{0.01\linewidth} p{0.06\linewidth} p{0.06\linewidth} p{0.01\linewidth}}
    \begin{tabular}{c c c c c c c c c c}
        \toprule
            &\multicolumn{3}{c}{Fixed-CNN} &\multicolumn{3}{c}{$+$ Puppeteer-Puppet} &\multicolumn{3}{c}{$+$ Parameter Adaptation} \\
        \cmidrule(lr){2-4}\cmidrule(lr){5-7}\cmidrule(lr){8-10}
            &Top-1 &Top-5 &Params &Top-1 &Top-5 &Params &Top-1 &Top-5 &Params \\
        \midrule
            AlexNet &65.86$\%$ &92.05$\%$ &11.65 &69.33$\%$ &94.05$\%$ &1.08 &70.05$\%$ &94.13$\%$ &1.08 \\ 
            VGG &67.84$\%$ &91.98$\%$ &39.01 &67.78$\%$ &93.02$\%$ &1.08 &68.42$\%$ &90.13$\%$ &1.08 \\ 
            ResNet &68.94$\%$ &95.17$\%$ &44.84 &66.69$\%$ &95.46$\%$ &1.08 &68.25$\%$ &94.31$\%$ &1.08 \\
         \bottomrule
    \end{tabular}
    %\vspace{-0.2in}
\end{table*}
Table.~\ref{tab:compression_cifar_cls} compares the original backbones with their Puppet variants.
Across these architectures, replacing independently learned layer parameters with parameters generated from a shared evolution process maintains competitive predictive performance while dramatically reducing the number of trainable parameters.
This suggests that continuous parameter evolution can serve as a viable alternative to conventional layer-wise parameterization across diverse CNN structures.
\begin{table*}[!htpb]
    %\vspace{-0.2in}
    \centering
    \renewcommand\arraystretch{0.8}	
    %\small
    \caption{Comparison between Puppet-CNN and lightweight CNN architectures.}% on Cifar-10.}
    \label{tab:scheme_cifar_cls}
    %\vspace{-0.1in}
    \begin{tabular}{c c c c c}
         \toprule
         &Top-1 Acc &Top-5 Acc &Params &Speed  \\
         \midrule
         AlexNet &65.86$\%$ &92.05$\%$ &11.65 &0.0006 \\
         MobileNet-v1 &58.58$\%$ &91.17$\%$ &7.52 &0.0018 \\ 
         MobileNet-v2 &66.73$\%$ &93.33$\%$ &8.90 &0.0032 \\ 
         SqueezeNet &63.31$\%$ &94.34$\%$ &3.96 &0.0022 \\ 
         \midrule
         \textbf{Puppet-CNN} &\textbf{72.51$\%$} &\textbf{96.85$\%$} &\textbf{1.08} &0.0039 \\ 
         \bottomrule
    \end{tabular}
    %\vspace{-0.1in}
\end{table*}

To further contextualize the efficiency and performance characteristics of the proposed model, Table.~\ref{tab:scheme_cifar_cls} compares Puppet-CNN with several lightweight CNN architectures designed for parameter-efficient learning. Under the same experimental setting, Puppet-CNN achieves competitive accuracy while using substantially fewer parameters, highlighting the parameter efficiency naturally induced by the proposed parameter evolution mechanism.
\begin{table}[!htpb]
    %%\vspace{-0.2in}
    \centering
    \renewcommand\arraystretch{0.8}	
    %\small
    \caption{Top-1 / Top-5 accuracy comparison between fixed CNNs and their Puppet-Puppeteer variants under partial (10k samples) and full (complete training set) CIFAR-10 training data.}
    \label{tab:partial_full}
    %\vspace{-0.1in}
    \begin{tabular}{c c c c c}
        \toprule
             &\multicolumn{2}{c}{Partial Set} &\multicolumn{2}{c}{Full Set} \\
        \cmidrule(lr){2-3}\cmidrule(lr){4-5}
            & Fixed-CNN & Puppet-Puppeteer & Fixed-CNN & Puppet-Puppeteer \\
        \midrule
            AlexNet &65.86$\%$/92.05$\%$ &69.33$\%$/94.05$\%$ &80.89$\%$/93.91$\%$ &83.75$\%$/96.90$\%$ \\ 
            VGG &67.84$\%$/91.98$\%$ &67.78$\%$/93.02$\%$ &81.05$\%$/95.37$\%$ &80.57$\%$/95.13$\%$ \\
            ResNet &68.94$\%$/95.17$\%$ &66.69$\%$/95.46$\%$ &82.24$\%$/96.11$\%$ &80.47$\%$/96.70$\%$ \\
        \bottomrule
    \end{tabular}
    %\vspace{-0.2in}
\end{table}

For reference, we also report the corresponding accuracy comparison under both partial and full training sets in Table.~\ref{tab:partial_full}. Across different backbone templates, the Puppet variants remain comparable to their fixed counterparts in both regimes, indicating that the effectiveness of the proposed parameterization is not limited to a particular training data scale.

\noindent\textbf{Parameter-Level Adaptation.}
Since all convolutional parameters in the puppet module are generated through the shared parameter evolution process, the framework naturally enables input-adaptive parameter generation conditioned on input complexity. As shown in Table.~\ref{tab:compression_cifar_cls}, introducing parameter adaptation improves Top-1 accuracy over the Puppet-Puppeteer-only variant in some backbones, and helps close the gap to the corresponding fixed-parameter baselines.

\noindent\textbf{Depth-Level Adaptation.}
Generating convolutional parameters through the puppeteer ODE module introduces additional computational cost when the network structure is kept fixed. To control this overhead, our framework further introduces an input-dependent depth adaptation mechanism that adjusts the number of instantiated convolutional layers according to the input complexity. Table.~\ref{tab:adaptive_depth_cifar_cls} compares three variants: the original ResNet with fixed parameters, a Puppet-ResNet where parameters are generated but the network depth remains fixed, and the proposed Puppet-CNN with adaptive depth. While Puppet-ResNet incurs substantially higher computational cost due to parameter generation, Puppet-CNN maintains a number of operations close to that of the original ResNet. This observation suggests that the proposed depth adaptation mechanism provides an effective way to regulate the computational cost of the parameter evolution process. Together with the results in Table.~\ref{tab:compression_cifar_cls}, these results also support the effectiveness of the input complexity measurement introduced in Sec.~\ref{sec:3.2}.
\begin{table}[!htpb]
    %\vspace{-0.2in}
    \centering
    \renewcommand\arraystretch{0.8}	
    %\small
    \caption{Computational cost comparison between fixed-depth and adaptive-depth variants.}%on Cifar-10.}
    \label{tab:adaptive_depth_cifar_cls}
    %\vspace{-0.1in}
    \begin{tabular}{c c c c}
         \toprule
        &Params &Mult-Adds(G) &Speed \\
         \midrule
            ResNet &44.84 &11.40 &0.0019 \\
            Puppet-ResNet &1.08 &23.44 &0.0061 \\
         \midrule   
            \textbf{Puppet-CNN} & 1.08 &12.34 &0.0039 \\
         \bottomrule
    \end{tabular}
    %\vspace{-0.2in}
\end{table}

\subsection{Robustness Study~\label{sec:4.5}}

To further examine the generalization of the proposed parameter evolution mechanism, we evaluate Puppet-CNN on more challenging classification datasets, including CIFAR-100 and mini-ImageNet. Compared with CIFAR-10, these datasets contain more categories and fewer training samples per category, resulting in a more challenging classification setting. In our CIFAR-10 experiments, only a subset of the training data is used, resulting in approximately 800 training samples per category. In CIFAR-100, each class contains only 500 images in total, with 400 used for training. A similar situation also appears in mini-ImageNet, where the number of training samples per category is also limited. Table.~\ref{tab:overall_cifar_100_cls} reports the performance comparison between Puppet-CNN and several representative CNN backbones on these datasets. We focus on relative behavior under a unified training protocol rather than saturating benchmark accuracy. Across both datasets, Puppet-CNN maintains competitive Top-1 and Top-5 accuracy while using substantially fewer parameters. These results suggest that organizing convolutional parameters through continuous evolution remains effective in more challenging classification scenarios and can provide a compact parameterization of convolutional networks. Similar trends across different datasets indicate that the proposed parameter evolution mechanism generalizes beyond a specific dataset or training configuration.
\begin{table}[!htpb]
    %\vspace{-0.2in}
    \centering
    \renewcommand\arraystretch{0.8}	
    %\small
    \caption{Performance comparison on CIFAR-100 and mini-ImageNet.}
    \label{tab:overall_cifar_100_cls}
    %\vspace{-0.1in}
    \begin{tabular}{c c c c c c c c}
        \toprule
            &\multicolumn{3}{c}{CIFAR-100} &\multicolumn{3}{c}{mini-ImageNet}  \\
        \cmidrule(lr){2-4}\cmidrule(lr){5-7}
             &Top-1 Acc &Top-5 Acc &Params &Top-1 Acc &Top-5 Acc &Params   \\
        \midrule
            AlexNet &41.46$\%$ &67.67$\%$ &11.84 &36.92$\%$ &64.35$\%$ &11.84 \\
            VGG &45.06$\%$ &67.66$\%$ & 39.20 &32.95$\%$ &59.94$\%$ & 39.20 \\
            ResNet &44.07$\%$ &70.96$\%$ &45.02 &38.41$\%$ &68.58$\%$ &45.02 \\
            \textbf{Puppet-CNN} &\textbf{53.26$\%$} &\textbf{79.63$\%$} &\textbf{1.08} &\textbf{46.46$\%$} &\textbf{74.38$\%$} &\textbf{1.08} \\
        \bottomrule
    \end{tabular}
    %\vspace{-0.2in}
\end{table}

\subsection{Quantitative Study~\label{app:a}}

\begin{figure*}[!htpb]
    %%\vspace{-0.2in}
    \centering
    \begin{subfigure}{.48\linewidth}
        \centering
        \includegraphics[width=\linewidth]{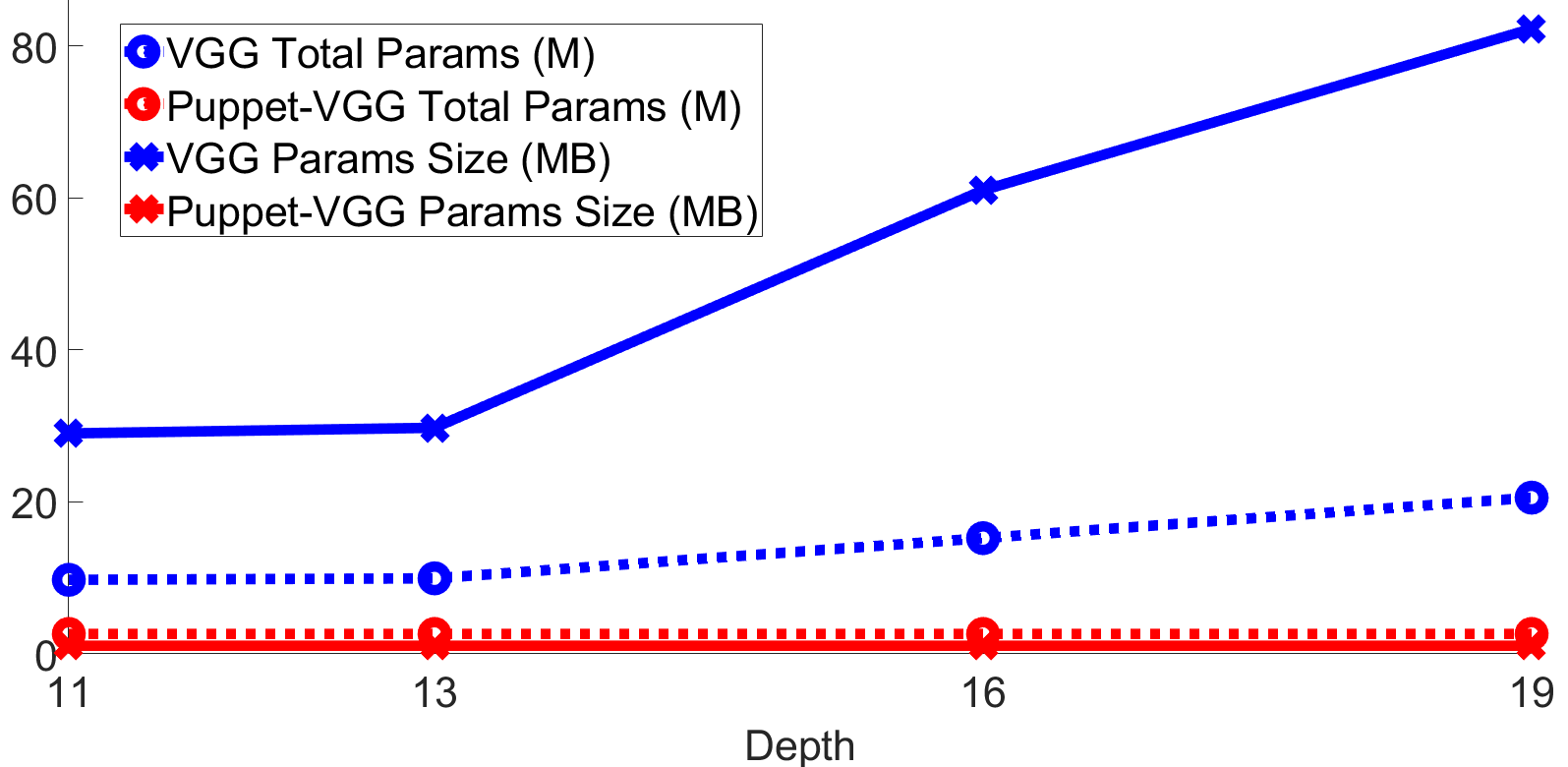}
        \subcaption[]{VGG vs. Puppet-VGG}
        \label{fig:depth_param_vgg}
    \end{subfigure}
    \hfill
    \begin{subfigure}{.48\linewidth}
        \centering
        \includegraphics[width=\linewidth]{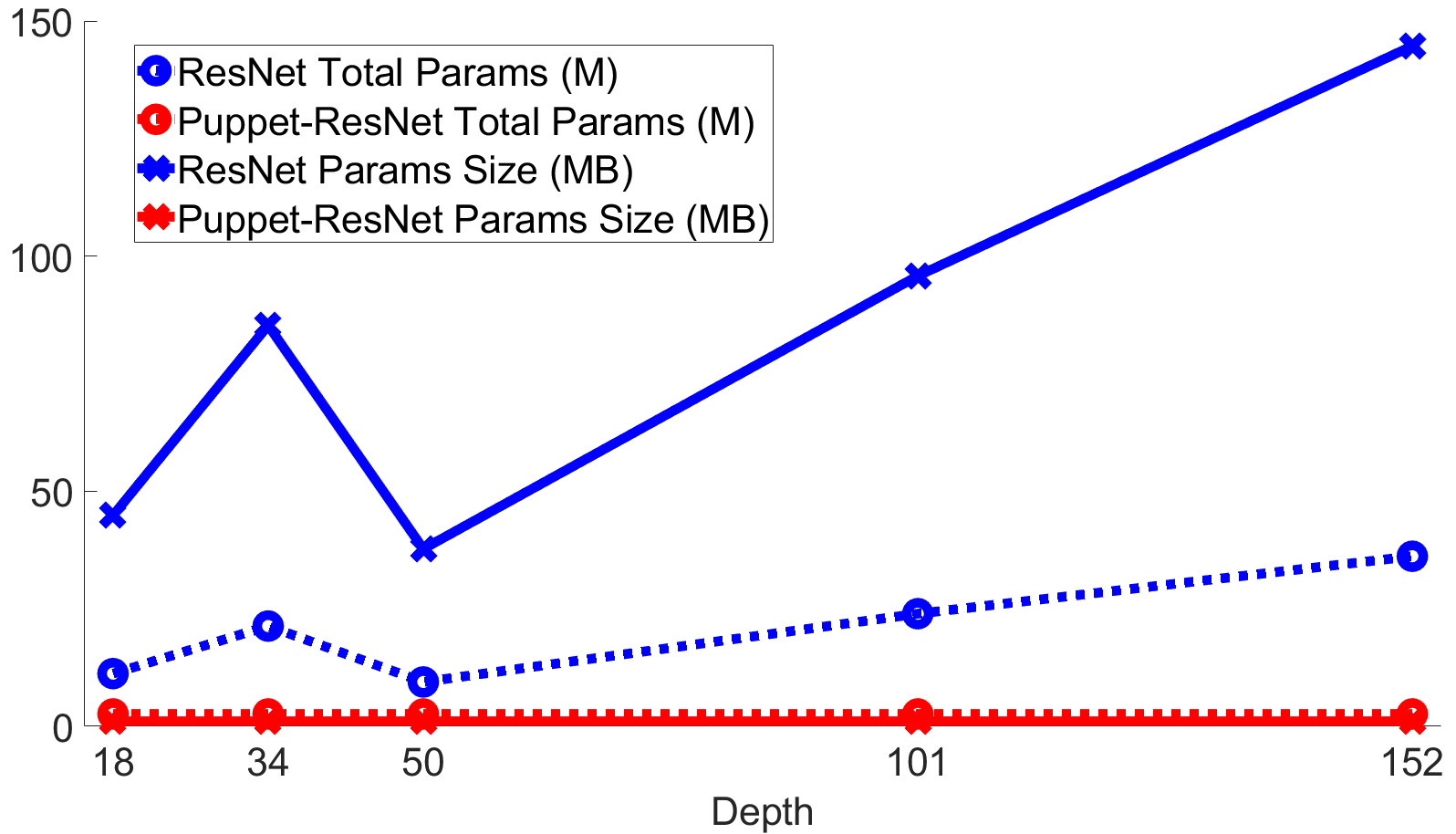}
        \subcaption[]{ResNet vs. Puppet-ResNet}
        \label{fig:depth_param_resnet}
    \end{subfigure}
    %\vspace{-0.1in}
    \caption{Model size comparison between conventional CNNs and their Puppet counterparts under increasing network depth for VGG and ResNet architectures.}
    \label{fig:depth_param}
    %%\vspace{-0.2in}
\end{figure*}

We further analyze the parameterization behavior of the proposed Puppet-CNN with respect to network depth and channel capacity. Fig.~\ref{fig:depth_param} illustrates how model size changes as network depth increases for VGG and ResNet architectures. In this experiment, the depth-adaptation component is disabled so that the depth of the puppet module can be manually controlled. The puppet module follows the architectural templates of VGG and ResNet while the number of layers is increased.
For conventional CNNs, the number of parameters grows with network depth because each layer contains independently learned convolutional kernels. In contrast, the Puppet variants maintain nearly constant parameter sizes across different depths since convolutional kernels are generated from the shared parameter evolution process. This behavior indicates that the proposed parameterization effectively decouples model size from network depth.
\begin{table*}[!htpb]
    %%\vspace{-0.2in}
    \centering
    \renewcommand\arraystretch{0.8}	
    %\small
    \caption{Parameter count and model size under different maximum output channel settings in the puppet module.}% on Cifar-10.}
    \label{tab:output_channel_param}
    %\vspace{-0.1in}
    \begin{tabular}{c c c c c c c c}
        \toprule
            $C_{\text{max}}^{\text{out}}$ &64 &128 &256 &512 &1024 &2048 &4096 \\
        \midrule
            Total Params (M) &0.04 &0.17 &0.66 &2.63 &10.50 &41.97 &167.83 \\
            Param Size (MB) &0.02 &0.07 &0.28 &1.08 &4.26 &16.90 &67.35 \\
        \bottomrule
    \end{tabular}
    %\vspace{-0.2in}
\end{table*}
We also examine how the parameter size changes with respect to the maximum number of output channels in the puppet module. As described in Sec.~\ref{sec:3.3}, the number of variables in the puppeteer ODE module depends on the maximum output channel capacity. Table.~\ref{tab:output_channel_param} reports the resulting parameter counts under different channel configurations. As expected, increasing the maximum output channel leads to a quadratic growth in the number of parameters due to the dimensionality of the generated convolutional kernels. Nevertheless, the overall parameter size remains relatively small compared to conventional CNN models with independently learned parameters.

%--------------------------------------------------
%       Section 5 : Conclusion
%--------------------------------------------------

%% 1/2 Page Max
\vspace{-0.1in}
\section{Conclusion~\label{sec:5}}

We propose Puppet-CNN, a convolutional framework in which convolutional kernels and network depth are generated through a shared parameter evolution process. In the proposed architecture, the parameters of the convolutional layers are produced by a puppeteer ODE module conditioned on the input complexity, enabling sample-dependent parameter and depth adaptation. The proposed parameterization significantly reduces the number of trainable parameters while allowing deep convolutional architectures to be instantiated through a compact evolution mechanism. In addition, the framework is compatible with existing CNN backbones and can be integrated into different network structures. Experimental results across several image classification benchmarks demonstrate that Puppet-CNN maintains competitive predictive performance while using substantially fewer parameters than conventional CNN architectures. These results suggest that organizing convolutional parameters through continuous evolution provides an effective alternative to conventional layer-wise parameterization. In future work, we plan to explore the application of this parameter evolution mechanism to additional vision tasks and investigate its extension to other neural network architectures beyond convolutional models. Improving the efficiency of the parameter generation process is another potential direction for further study.

%Bibliography
\bibliographystyle{splncs04}
\bibliography{eccv26}

\end{document}